\pdfoutput=1

\documentclass[11pt]{article}
\usepackage[dvipsnames,table,xcdraw,svgnames]{xcolor}
\usepackage[final]{acl} 
\usepackage{times} 
\usepackage{latexsym} 
\usepackage{microtype} 
\usepackage{booktabs} 
\usepackage{multirow} 
\usepackage{graphicx} 
\usepackage{amsmath} 
\usepackage{amsmath}  
\usepackage{amssymb}  
\usepackage{bm} 
\usepackage{caption} 
\usepackage{subfigure} 
\usepackage{algorithm2e} 
\usepackage{algpseudocode} 
\usepackage{mdwlist} 
\usepackage{url} 
\usepackage{inconsolata} 
\usepackage{pifont} 
\usepackage{mathptmx}  

\usepackage{color} 
\usepackage{makecell} 
\usepackage[misc]{ifsym} 
\usepackage[utf8]{inputenc} 
\usepackage[bottom]{footmisc} 
\usepackage{CJKutf8}  
\usepackage{longtable}
\usepackage{supertabular}

\usepackage{float} 
\usepackage{geometry}
\usepackage{tablefootnote}
\usepackage{threeparttable}
\usepackage{ulem}
\usepackage{hyperref} 
\usepackage{ulem} 
\definecolor{Gray}{gray}{0.93}
\definecolor{Red}{RGB}{255, 46, 23}
\usepackage{fontawesome}
\usepackage{amsfonts}   
\usepackage{amssymb}     
\usepackage{mathptmx}    

\usepackage{hyperref}
\usepackage{etoolbox}
\appto\frontmatter{\hypersetup{bookmarksdepth=-1}}  

\title{Making LLMs Better Many-to-Many \\Speech-to-Text Translators with Curriculum Learning}

\author{Yexing Du$^{1,2}$ \quad Youcheng Pan$^{2}$ \quad Ziyang Ma$^{3}$ \quad Bo Yang$^{2}$ \quad Yifan Yang$^{3}$ \\ \textbf{Keqi Deng$^{4}$ 
\quad Xie Chen$^{3}$ \quad Yang Xiang$^{2}$$^{*}$ \quad Ming Liu$^{1,2}$\thanks{ Corresponding author.} \quad Bing Qin$^{1,2}$} \\
$^1$Harbin Institute of Technology \quad $^2$Pengcheng Laboratory\\ \quad $^3$Shanghai Jiao Tong University \quad $^4$University of Cambridge \\ 
\texttt{{\{yxdu, mliu\}@ir.hit.edu.cn,}  {\{panych,xiangy\}@pcl.ac.cn}}
}

\begin{document}
\maketitle

\makeatletter
\def\thanks#1{\protected@xdef\@thanks{\@thanks
        \protect\footnotetext{#1}}}
\makeatother


\begin{abstract}
Multimodal Large Language Models (MLLMs) have achieved significant success in Speech-to-Text Translation (S2TT) tasks. While most existing research has focused on English-centric translation directions, the exploration of many-to-many translation is still limited by the scarcity of parallel data. To address this, we propose a three-stage curriculum learning strategy that leverages the machine translation capabilities of large language models and adapts them to S2TT tasks, enabling effective learning in low-resource settings. We trained MLLMs with varying parameter sizes (3B, 7B, and 32B) and evaluated the proposed strategy using the FLEURS and CoVoST-2 datasets. Experimental results show that the proposed strategy achieves state-of-the-art average performance in $15\times14$ language pairs, requiring fewer than 10 hours of speech data per language to achieve competitive results. \footnote{The source code and models are released at \url{https://github.com/yxduir/LLM-SRT}.}

\end{abstract}
\section{Introduction}

\begin{figure}[t]
 \centering
    \setlength{\belowcaptionskip}{-0.25cm} 
    \includegraphics[scale=0.9]{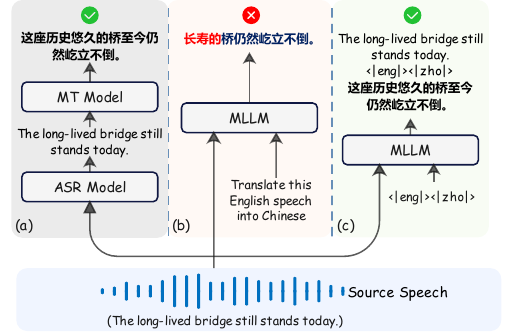}
    \caption{\textbf{Comparison of S2TT Methods.} (a) adopts a cascaded system; (b) directly generates translated text; (c) generates both transcription and translation text in an end-to-end process, with \texttt{<|eng|><|zho|>} indicating transcribing English and translating it into Chinese.}
    \label{example1}
\end{figure}

Speech-to-Text Translation (S2TT) involves converting speech from a source language into text in a target language. Traditionally, S2TT tasks have relied on a cascaded system, as shown in Figure \ref{example1}(a), where an Automatic Speech Recognition (ASR) module transcribes speech into text~\cite{baevski2020wav2vec, gulati2020conformer}, followed by a Machine Translation (MT) module that translates the transcribed text into the target language~\cite{cheng2019breaking, beck2019neural}. However, this cascade system often suffers from error propagation~\cite{sperber2020speech}.  Recently, Multimodal Large Language Models (MLLMs), illustrated in Figure \ref{example1}(b), have demonstrated advantages in simplifying model architecture and mitigating error propagation in both ASR~\cite{zhang2023speechgpt, ma2024embarrassingly} and S2TT tasks~\cite{chu2024qwen2}.

Current MLLMs process \{\textit{speech}, \textit{instruction}\} inputs to directly generate \{\textit{translation}\}, but this approach heavily relies on large-scale S2TT datasets. Existing datasets~\cite{wang2020covost, di2019must} predominantly focus on English, while datasets supporting many-to-many S2TT, such as FLEURS~\cite{fleurs2022arxiv}, remain limited. Meanwhile, Large Language Models (LLMs) have demonstrated strong many-to-many multilingual MT capabilities. This raises the question: can the MT capabilities of LLMs be effectively transferred to the S2TT task with limited data?

Inspired by advances in transfer learning~\cite{pham2024unibridge,mueller2024multi}, we transform the S2TT task into a \textbf{Speech Recognition and Translation (SRT)} task, which  involves training \{\textit{speech}, \textit{instruction}\} to generate \{\textit{transcription}, \textit{translation}\}, as shown in Figure~\ref{example1}(c). 
LLMs possess robust MT capabilities, which can be adapted to S2TT tasks with minimal data. This approach allows MLLMs to harness MT capabilities for many-to-many S2TT, effectively combining the advantages of both cascade and end-to-end models.

To connect MT and S2TT tasks, we propose a three-stage curriculum learning strategy: (1) \textbf{ASR}, which trains the MLLM for multimodal alignment, enabling the model to understand speech and generate transcriptions; (2) \textbf{Speech-Aided Machine Translation (SMT)}, where both speech and transcription are provided, and the MLLM generates translations to improve cross-lingual capabilities; (3) \textbf{SRT}, where only speech is provided, and the MLLM generates both transcription and translation. The training proceeds sequentially through these three stages, with each stage resuming from the checkpoint of the previous one. The resulting MLLM achieves many-to-many S2TT through the SRT task. Additionally, we designed specialized instructions for many-to-many S2TT and implemented an optimized lightweight speech adapter for efficient speech feature compression to accelerate inference.

To evaluate our strategy, we trained three MLLM variants (3B, 7B, and 32B). In low-resource scenarios (with fewer than 10 hours of data per language), our MLLM, trained on the FLEURS dataset, demonstrated strong many-to-many S2TT capabilities, outperforming existing state-of-the-art end-to-end models. We also assessed performance on the EN-X direction of the CoVoST-2 dataset, where sufficient training data is available, and found that our strategy remains effective, surpassing state-of-the-art models.

The key contributions of this work are:

\begin{itemize}
    \item This paper adopts a strategy that transforms the S2TT task into an SRT task, leveraging the machine translation capabilities of LLMs to enhance the many-to-many S2TT performance of MLLMs, particularly in low-resource settings.
    
    \item We propose a three-stage curriculum learning strategy and systematically evaluate our strategy across datasets of varying scales and model sizes (3B, 7B, 32B). To the best of our knowledge, our model is the first MLLM to support many-to-many S2TT at the 32B scale.

    \item Our model achieves state-of-the-art average performance across 15×14 translation directions in the S2TT task under low-resource settings on the FLEURS dataset, while also demonstrating strong robustness on the CoVoST-2 dataset in high-resource scenarios.
\end{itemize}


\begin{figure*}[t]
\centering
    \includegraphics[scale=1]{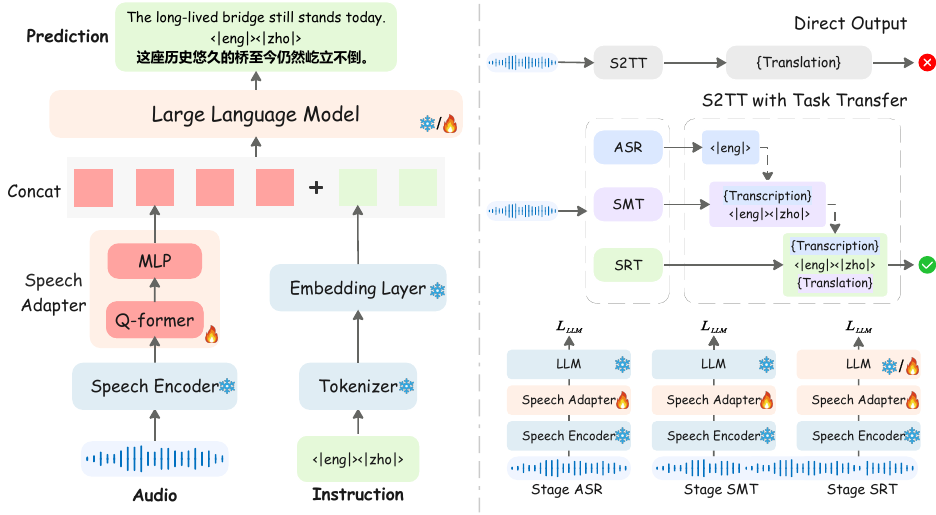}
    \caption{\textbf{The Architecture of LLM-SRT.} LLM-SRT consists of a speech encoder, speech adapter, and LLM. A three-stage curriculum learning strategy sequentially trains the ASR, SMT, and SRT tasks, as shown in Table \ref{pattern}. In stages 1 and 2, the speech adapter is continuously trained to enable efficient fine-tuning. In stage 3, the LLM is additionally unfrozen, while the speech adapter continues to be trained.}
    \label{framework}
\end{figure*}
\section{Methodology}
In this section, we present the methodology of our approach. Section \ref{sec:3.0} defines the tasks involved in our method. Section \ref{sec:3.1} presents the architecture of the LLM-SRT model. Section \ref{sec:3.2} explains our curriculum learning strategy, which sequentially fine-tunes the model for ASR, SMT, and SRT tasks.

\subsection{Problem Formulation}
\label{sec:3.0}
In this section, we define the following tasks: 

\textbf{ASR}: Given the audio input \( \mathbf{X} \) and the instruction text \( \mathbf{T} \), the goal is to produce the transcribed text \( \mathbf{Y} \).

\textbf{SMT}: Given the audio input \( \mathbf{X} \), its corresponding transcription \( \mathbf{Y} \), and the instruction text \( \mathbf{T} \), the goal is to produce the translated text \( \mathbf{Z} \).

\textbf{SRT}: Given the audio input \( \mathbf{X} \) and the instruction text \( \mathbf{T} \), the goal is to produce both the transcription \( \mathbf{Y} \) and the translated text \( \mathbf{Z} \).

\subsection{LLM-SRT}
\label{sec:3.1}

\begingroup
\renewcommand{\arraystretch}{1.25} 
\begin{CJK}{UTF8}{gbsn}
\begin{table*}[t]
  \centering
  \small
  \setlength{\tabcolsep}{6pt} 
    \begin{tabular}{c c c c} 
    \toprule 

\textbf{Task}&\textbf{Audio}&\textbf{Instruction}&\textbf{Prediction} \\ 
\hline

ASR &{\textcolor{green}{\ding{51}}}& \texttt{<|eng|>}  & Will it rain tomorrow? \\  
ASR &{\textcolor{green}{\ding{51}}}& \texttt{<|zho|>}  & 明天会下雨吗？ \\  \hline

SMT &{\textcolor{green}{\ding{51}}}& Will it rain tomorrow?\texttt{<|eng|><|deu|>}   & Regnet es morgen?    \\ 
SMT &{\textcolor{green}{\ding{51}}}& 明天会下雨吗？\texttt{<|zho|><|jpn|>}  & 明日は、雨かな？  \\ \hline

SRT &{\textcolor{green}{\ding{51}}}& \texttt{<|eng|><|deu|>} & Will it rain tomorrow?\texttt{<|eng|><|deu|>}Regnet es morgen?   \\ 
SRT &{\textcolor{green}{\ding{51}}}& \texttt{<|zho|><|jpn|>}  & 明天会下雨吗？\texttt{<|zho|><|jpn|>}明日は、雨かな？  \\

\bottomrule
    \end{tabular}
\caption{\textbf{Instruction Design.} The instruction design is intended for fine-tuning instructions for three tasks: ASR, SMT, and SRT, using simple yet effective instructions to distinguish between them.}
\label{pattern}
\end{table*}%
\end{CJK}
\endgroup

The LLM-SRT architecture is shown in Figure~\ref{framework}. The speech encoder extracts features from the speech input, and the speech adapter layer connects these features to the LLM, aligning their dimensions and incorporating speech feature compression. Finally, the LLM generates textual output by processing the concatenated embeddings derived from both text and speech features.

\paragraph{Speech Encoder.} The speech encoder processes the audio input \( \mathbf{X} \) into a high-dimensional representation using the frozen Whisper encoder~\cite{radford2023robust}, which has been pretrained on large-scale supervised datasets for speech recognition and translation.
\begin{equation}
\mathbf{H} = \text{Encoder}(\mathbf{X}),
\end{equation}
where \( \mathbf{H} \in \mathbb{R}^{T \times D} \) is the encoder's output, with \( T \) representing the time dimension and \( D \) the hidden dimension of the encoder.

\paragraph{Speech Adapter.} The speech adapter compresses the time dimension \( T \) and adjusts the hidden dimension \( D \) to match the LLM's hidden dimension \( d_{\text{LLM}} \).

We use a Q-Former to convert input sequences into fixed-length query representations.
\begin{equation}
\mathbf{Q'} = \text{Q-Former}(\mathbf{Q}, \mathbf{H}),
\end{equation}
where \( \mathbf{Q'} \in \mathbb{R}^{n_q \times D_q} \) is the output of the Q-Former, \( \mathbf{Q} \in \mathbb{R}^{n_q \times D_q} \) is the trainable query matrix, \( n_q \) is the number of queries, and \( D_q \) is the hidden dimension of the Q-Former.

After the Q-Former layer, a multilayer perceptron (MLP) projects the feature dimensions from \( D_q \) to \( d_{\text{LLM}} \):
\begin{equation}
\mathbf{E^X} = \text{ReLU}(\mathbf{Q'} \mathbf{W}_1 + \mathbf{b}_1) \mathbf{W}_2 + \mathbf{b}_2,
\end{equation}
where \( \mathbf{E^X} \in \mathbb{R}^{n_q \times d_{\text{LLM}}} \) is the output of the MLP layer, ready for LLM processing.

\paragraph{Tokenizer and LLM.} The tokenizer and embedding layer process the instruction \( \mathbf{T} \) and produce \( \mathbf{E^T} \in \mathbb{R}^{n_t \times d_{\text{LLM}}} \), where \( n_t \) is the length of the text tokens.

The speech features \( \mathbf{E^X} \) and text features \( \mathbf{E^T} \) are concatenated and fed into the frozen LLM:  
\begin{equation}
\mathbf{E^Z} = \mathbf{E^X} \oplus \mathbf{E^T},
\end{equation}  
where \( \mathbf{E^Z} \in \mathbb{R}^{(n_t + n_q) \times d_{\text{LLM}}} \) is processed by the LLM to generate the output text.

The output text varies depending on the task. During training, the parameters of the adapter layer are updated based on the loss of the LLM's output.

\subsection{Curriculum Learning}
\label{sec:3.2}

LLM-SRT adopts a curriculum learning approach that incorporates three training tasks: ASR, SMT, and SRT.

\paragraph{Instruction Design.} 
We designed minimalist instructions to help the model distinguish between tasks while reducing the instruction token length, as shown in Table~\ref{pattern}. This design ensures that instructions like \texttt{<|eng|><|deu|>} appear in the generated answers, effectively segmenting transcription and translation content in the SRT task.
\paragraph{ASR.} In this stage, the model is pre-trained to develop ASR capabilities with a focus on multimodal alignment, while expanding language support by training on all intended languages. The speech adapter is trained with as much data as possible to ensure efficient fine-tuning.

\paragraph{SMT.} This stage enhances the model's cross-lingual abilities. Starting from the ASR checkpoint, the model takes both transcribed text and audio as input to generate translations based on the instruction. The purpose of this step is to activate the LLM's inherent machine translation capabilities and establish the connection between the MT and S2TT tasks.

\paragraph{SRT.} This stage activates the SRT capabilities of the MLLM, finalizing the model. Training continues from the SMT checkpoint, with the model receiving only audio input and a task-specific instruction, outputting both the transcription and translation of the speech. This extends the MT capabilities of LLMs to the S2TT task.

\section{Experiments}

\subsection{Datasets}

\paragraph{FLEURS.} FLEURS\footnote{\url{https://huggingface.co/datasets/google/fleurs}}~\cite{fleurs2022arxiv} serves as the speech counterpart to the FLoRes\footnote{\url{https://huggingface.co/datasets/facebook/flores}}~\cite{costa2022no} machine translation benchmarks. It includes 102 languages, with each training set containing approximately 10 hours of supervised speech data per language.

\paragraph{CoVoST-2.} CoVoST-2\footnote{\url{https://github.com/facebookresearch/covost}}~\cite{wang2020covost} is a large-scale multilingual S2TT corpus derived from the Common Voice dataset~\cite{commonvoice:2020}. It contains translations from 21 languages to English and from English to 15 other languages.

\subsection{Experiment Settings}

\paragraph{Model Architecture.} The baseline model consists of an LLM (Qwen 3B, 7B, 32B), a frozen speech encoder (Whisper-large-v3), and a trainable adapter layer comprising a Q-Former and an MLP. Following the configuration in \citet{yu2024connecting}, we use 80 queries, each with a dimension of 768. Training can be minimized by freezing the LLM, or LoRA \cite{hu2021lora} can be applied for training.

\paragraph{Training Details.}
We used bf16 precision with Distributed Data Parallel (DDP), a learning rate of \(1 \times 10^{-4}\), 1000 warmup steps, and the AdamW optimizer. The models were trained on four A100 GPUs. We provide detailed settings in Table \ref{tab:details}, \ref{tab:models}, \ref{tab:lora}, and \ref{tab:train} of the Appendix.

For the ASR task, we used the Common Voice\footnote{\url{https://commonvoice.mozilla.org/en/datasets}} dataset, and for the SMT and SRT tasks, we used the FLEURS and CoVoST-2 datasets. In a 4-card A100 environment, the 3B and 7B models train in 3 days, while the 32B model trains in 7 days.

\subsection{Compared Methods}
We compare both cascade and end-to-end S2TT models, all of which support many-to-many S2TT.

\begin{itemize}
\item \textbf{Cascaded systems} are pipeline-based approaches, where an ASR model first transcribes speech into text, which is then translated by a machine translation model.

\item \textbf{SeamlessM4T}\footnote{\url{https://github.com/facebookresearch/seamless_communication}}~\cite{barrault2023seamlessm4t} is a foundational multilingual and multitask model capable of seamlessly translating and transcribing both speech and text. It supports speech-to-text translation for nearly 100 input and output languages.

\item \textbf{Qwen-Audio}\footnote{\url{https://github.com/QwenLM/Qwen-Audio}}~\cite{Qwen-Audio} is the multimodal extension of Qwen-LLM, designed to process diverse audio modalities, including speech, natural sounds, music, and songs, alongside text, generating text-based outputs.
\end{itemize}

\paragraph{Evaluation Metric.}
We use WER\footnote{\url{https://huggingface.co/spaces/evaluate-metric/wer}}~\cite{morris2004and}(for the ASR task) and BLEU\footnote{\url{https://github.com/mjpost/sacrebleu}}~\cite{post-2018-call} (for the S2TT task) as evaluation metrics.

\begin{table}[htbp]
  \centering
    \small
          \setlength{\tabcolsep}{3pt} 
          
 \begin{tabular}{cl} \toprule 
  \textbf{Metric}&\textbf{Details} \\ \midrule
\textbf{WER}& Text normalization follows Whisper\\ \cmidrule(r){2-2}
\multirow{4}{*}{\textbf{BLEU}}&  SacreBLEU signature:  \\
&\tiny nrefs:\texttt{1}\textbar case:\texttt{mixed}\textbar eff:\texttt{no}\textbar tok:\texttt{13a}\textbar smooth:\texttt{exp}\textbar version:\texttt{2.4.3}  \\
&Except for jpn, kor, tha, yue, zho with char: \\
&\tiny nrefs:\texttt{1}\textbar case:\texttt{mixed}\textbar eff:\texttt{no}\textbar tok:\texttt{char}\textbar smooth:\texttt{exp}\textbar version:\texttt{2.4.3}\\
\bottomrule
  \end{tabular}

  \caption{
\textbf{Metric Details.} We followed the settings of SeamlessM4T-V2~\cite{barrault2023seamlessm4t}.}
  \label{tab:metric}
\end{table}

\begin{table*}[htb]
  \centering
  \small
        \setlength{\tabcolsep}{6pt} 
\begin{tabular}{l|c|cccccc|c} \toprule 
    \multirow{2}{*}{X$\rightarrow$12 Languages}&\multirow{2}{*}{\shortstack{\textbf{S2TT Data} \\ (hour)}} &  \multicolumn{6}{c}{\textbf{FLEURS}}\\ 
    
      && Eng & Deu & Fra & Jpn & Rus & Zho & Avg. \\ \midrule       
            \multicolumn{9}{c}{Cascaded ASR+MT Methods}       \\ \midrule
    Whisper~+~Qwen2.5-3B&- & 23.4 & 21.0 & 20.9 & 13.6 & 19.6 & 14.5 & 18.8 \\
    Whisper~+~Qwen2.5-7B &-& 26.7 & 23.3 & 22.5 & 15.1 & 21.5 & 16.4 & 20.9 \\
    Whisper~+~Qwen2.5-32B &-& 29.9&26.0&24.6&17.3&23.8&18.5&23.3\\ \midrule 
     \multicolumn{9}{c}{End-to-End Models}       \\ \midrule
    SeamlessM4T-V2~(2.3B)&351,000& \uline{33.1} & \uline{20.5} & 19.6 & \uline{13.2} & \uline{19.6} & \uline{15.2} & \uline{20.2} \\
    
Qwen2-Audio~(7B)&in-house&22.6&20.1&\uline{20.6}&4.0&15.0&13.7&16.0 \\  
\rowcolor{Gray} {Baseline-3B} &52& 11.8 & 9.0 & 9.5 &5.2  &9.7&6.2&8.6  \\    
\rowcolor{Gray} {LLM-SRT-3B} &52& 27.2 & \textbf{22.6} & \textbf{22.0} & \textbf{14.3} & \textbf{21.3} & \textbf{16.5} & \textbf{20.6} \\
\rowcolor{Gray} {LLM-SRT-7B} &52& 27.4 & \textbf{23.7
}& \textbf{22.8} & \textbf{15.5} & \textbf{21.8} & \textbf{16.9} & \textbf{21.4} \\
\rowcolor{Gray} {LLM-SRT-32B} &52& 32.5 & \textbf{26.8} & \textbf{26.1} & \textbf{17.5} & \textbf{25.6} & \textbf{19.2} & \textbf{24.6} \\\bottomrule

\end{tabular}

  \caption{
\textbf{BLEU Scores on 6x12 Directions in FLEURS.} \uline{Underlined} denotes previous state-of-the-art end-to-end models, while \textbf{bold} indicates models that outperform them. "-" indicates no S2TT data was used due to the cascade system. The baseline uses the same instruction-tuning strategy as Qwen2-Audio.
}
  \label{tab:fleurs_6}
\end{table*}

\subsection{Overall Results}
As shown in Tables \ref{tab:fleurs_6} and \ref{tab:fleurs_15}, we evaluate the S2TT performance in low-resource settings on the FLEURS dataset. The results indicate that our model achieves state-of-the-art performance in the many-to-many S2TT task. Similarly, Table \ref{tab:covost2} presents the results under high-resource conditions on the CoVoST-2 dataset. Table \ref{tab:speed} provides a comparison of inference speed, and Table \ref{tab:ablation} presents an ablation study of the three-stage curriculum learning.

\paragraph{Language Support.}
As an LLM-based model designed for many-to-many S2TT, conducting comprehensive baseline comparisons is both essential and challenging. To ensure a thorough evaluation, we compare baselines across 6×12 language directions and benchmark our model against state-of-the-art approaches in the 15×14 setting. The supported languages are listed in Table \ref{tab:all_languages}, while the complete experimental results are provided in Table \ref{fleurs_15x14}.

\paragraph{Baseline-3B vs. LLM-SRT-3B.}

For the baseline model, we first conduct ASR pretraining as in Qwen2-Audio, followed by S2TT instruction-tuning with the same setup. Due to limited data, the baseline performs poorly, highlighting the limitations of traditional fine-tuning in low-resource settings.

In contrast, our curriculum learning training strategy achieves state-of-the-art performance (8.6$\rightarrow$20.6) in low-resource scenarios, as we transform S2TT into an SRT task, effectively leveraging the machine translation capability of the LLM to achieve many-to-many S2TT.

\paragraph{SeamlessM4T-V2 vs. LLM-SRT-3B.}

The LLM-SRT-3B demonstrates superior performance over SeamlessM4T-V2 on non-English languages (e.g., for French 18.8$\rightarrow$22.0), while it lags behind SeamlessM4T-V2 in English-to-X translation. This discrepancy can largely be attributed to the larger amount of S2TT data available for SeamlessM4T-V2, which includes 351,000 hours of training data compared to just 52 hours for LLM-SRT-3B. In situations where data resources are limited, our LLM-based approach has a greater advantage.

\paragraph{Cascaded Systems vs. LLM-SRT.}
As shown in Table \ref{tab:fleurs_6}, when using the same LLM as in the cascaded system, our MLLM demonstrates a clear performance advantage (e.g., for 3B, 18.8$\rightarrow$20.6), highlighting the benefits of the end-to-end approach. The MLLM's superior performance stems from its integrated framework, which eliminates intermediate steps and enables more efficient knowledge transfer, resulting in improved translation accuracy and robustness. This makes the model more effective in handling complex language tasks.

\begin{table*}[htb]
  \centering
  \scriptsize
      \setlength{\tabcolsep}{2pt} 
      \resizebox{\textwidth}{!}{ 
\begin{tabular}{l|c|ccccccccccccccc|cc} \toprule 
    \multirow{2}{*}{X$\rightarrow$14 Languages}&\multirow{2}{*}{\shortstack{\textbf{S2TT Data} \\ (hour)}} &  \multicolumn{15}{c}{\textbf{FLEURS}}\\ 
    
    &  & Eng & Deu & Fra& Ind& Ita & Jpn& Kor& Nld& Por & Rus& Spa & Tha& Vie& Yue & Zho & Avg. \\ \midrule   
           \multicolumn{18}{c}{Machine Translation}       \\ \midrule
{Qwen2.5-3B}&- & 29.5&25.4   & 25.2 & 24.6 & 22.7 & 18.4 & 18.8& 22.1& 27.1& 23.4&22.1& 18.3& 22.6& 19.0& 20.5& 22.6 \\ \midrule

           \multicolumn{18}{c}{Speech to Text Translation}       \\ \midrule

    SeamlessM4T-V2 (2.3B)&351,000& \uline{33.3} & 21.6 & 21.1 & 18.4 & 19.1 &14.5 & 17.2& 18.4& 18.8& 20.6& 17.6& 12.8& 17.5& 15.2& 16.7& 18.8 \\
    SeamlessM4T-V2 +Lora&351,129& 32.8 & 22.0 & 21.3 & 19.1 & 19.4 &14.4 & 17.0& 18.6& 18.6& 21.3& 18.1& 13.2& 17.5& 14.9& 17.0& 19.0 \\

\rowcolor{Gray}{LLM-SRT-3B-V2}&129 & 27.8 & 23.4 & 24.0 & 22.4 & 22.0 & 15.3 & 17.9& 20.4& 25.7& 22.8&21.9& 12.9& 18.2& 15.5& 19.2& 20.6 \\
\rowcolor{Gray}{LLM-SRT-3B-V2 +Lora}&129& 29.1 &\textbf{ 24.5 }&\textbf{ 24.3 }&\textbf{ 22.9 }&\textbf{ 22.8 }&\textbf{ 16.5 }&\textbf{ 18.0}&\textbf{ 20.9}&\textbf{ 26.2}&\textbf{ 23.3}&\textbf{22.6}&\textbf{ 14.7}&\textbf{ 19.0}&\textbf{ 16.3}&\textbf{ 19.6}&\textbf{ 21.4} \\\bottomrule 

\end{tabular}}
   \caption{
\textbf{BLEU Scores on 15×14 Directions.} The complete results are in Table \ref{tab:comet} and \ref{fleurs_15x14} in the Appendix.}
  \label{tab:fleurs_15}
\end{table*}
\paragraph{Scaling Law of LLM-SRT.}
As shown in Table 3, experiments on models with 3B, 7B, and 32B parameters demonstrate that our method follows the scaling law of LLMs (e.g., 20.6 for 3B, 21.4 for 7B, and 24.6 for 32B). Notably, the 32B model achieved state-of-the-art performance across all directions, confirming the generalizability of our approach. Our model’s performance is strongly correlated with the machine translation capability of the LLM, making the choice of an appropriate LLM foundation crucial for optimal performance.

\paragraph{Many-to-Many S2TT on FLEURS.} As shown in Tables \ref{tab:fleurs_15} and \ref{fleurs_15x14}, we compared performance across 15 languages and 210 translation directions. Table \ref{tab:fleurs_15} reports the average performance across the 15 languages, where our model achieves state-of-the-art BLEU performance (18.8$\rightarrow$21.4). Table 11 presents detailed results for all 210 directions, showing that our model outperforms SeamlessM4T-V2 in 154 directions.

\paragraph{Train Adapter Only vs. Fine-tune LLM.} As shown in Table \ref{tab:fleurs_15}, LLM-SRT-3B-V2 achieves high translation performance (20.6) by freezing the speech encoder and LLM, while training only the speech adapter. Further performance improvement (20.6$\rightarrow$21.4) can be achieved by unfreezing the LLM, such as through LoRA training.

\begin{figure}[htp]
    \vspace{-0.8cm} 
    \setlength{\abovecaptionskip}{-0.5cm} 
\centering
    \makebox[\textwidth][l]{\hspace{-0.4cm}\includegraphics[scale=0.28]{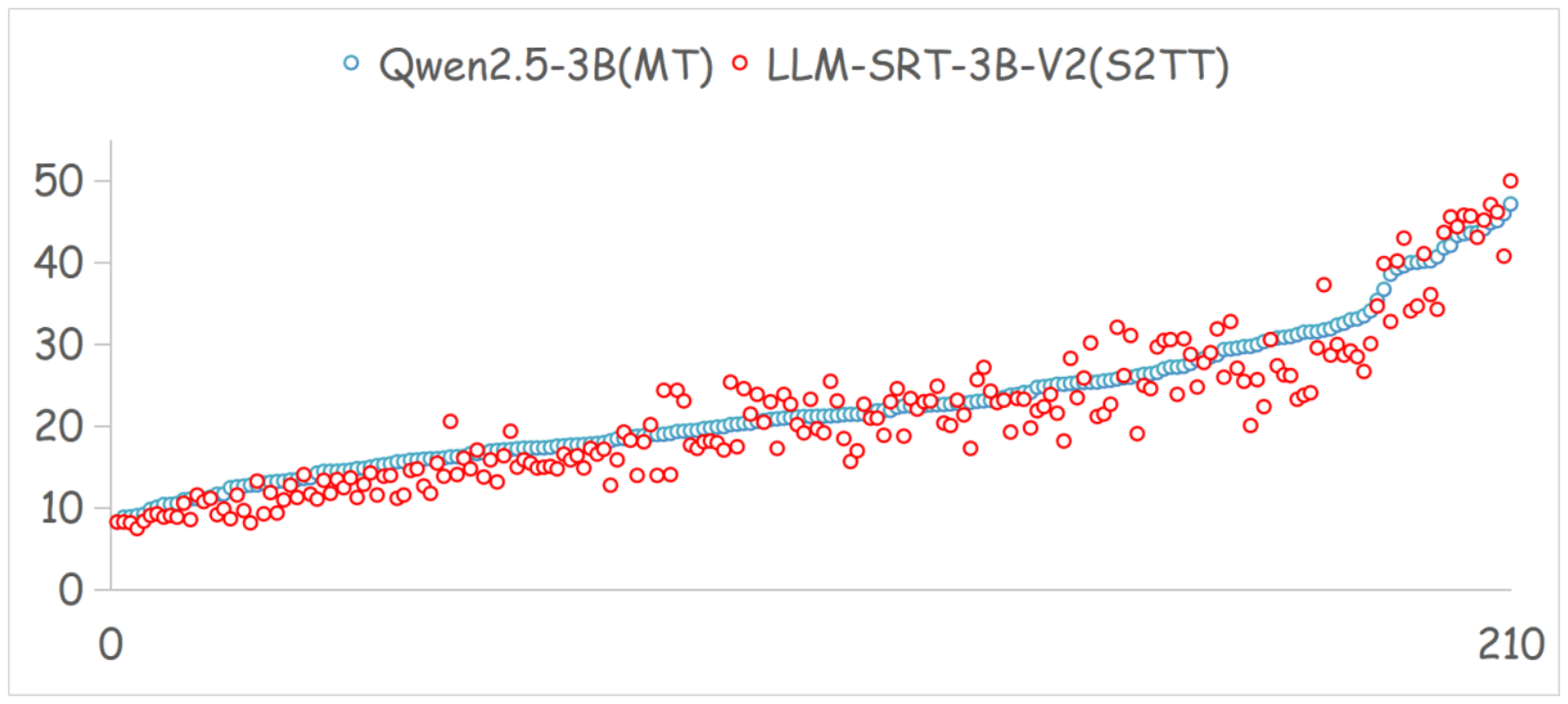}}
    \caption{\textbf{BLEU Scores for 15×14 Directions: Comparison between MT and S2TT.} The results show a strong correlation, suggesting that our S2TT capability is derived from the MT model. Table \ref{tab:error} includes an error analysis showing that S2TT outperforms MT. }

    \label{mt_st}
\end{figure}

\paragraph{MT vs. S2TT.} As shown in Figure \ref{mt_st}, we compared the MT performance of Qwen2.5-3B with the S2TT performance of LLM-SRT-3B-V2 across 210 translation directions. The results show a strong correlation between our MLLM’s S2TT and MT performance, confirming that the strong S2TT capability of LLM-SRT-3B stems from the LLM’s machine translation ability.

 \begin{table}[htb]
  \centering
    \small
          \setlength{\tabcolsep}{2pt} 
  \begin{tabular}{l|c|ccc|c} \toprule 
    \multirow{2}{*}{Eng$\rightarrow$X}& \multirow{2}{*}{\shortstack{\textbf{S2TT Data} \\ (hour)}} &  \multicolumn{3}{c}{\textbf{CoVoST-2}}\\ 
    
      && Deu& Jpn & Zho &Avg. \\ \midrule    
                \multicolumn{6}{c}{Cascaded ASR+MT Methods}       \\ \midrule
    {Whisper+NLLB-3.3B}&-&33.4&31.0     &  32.0 & 32.1 \\ 
    {Whisper+Qwen2.5-3B} &-&  21.1 &  28.1  &  35.0 & 28.0\\
        {Whisper+Qwen2.5-7B} &-&  24.1 &  29.8  &  37.6 & 30.5\\
    {Whisper+Qwen2.5-32B}&-&  28.3 &  35.0  &  41.1 & 34.8\\  \midrule
          \multicolumn{6}{c}{End-to-End Models}       \\ \midrule
    {SeamlessM4T-V2}&351,000&\uline{37.0}&\uline{39.7} &35.9  & 37.5 \\
      {Qwen-Audio~(7B)} &3,700&25.1&-   &40.3  &- \\ 
      {Qwen2-Audio~(7B)} &in-house&29.9&\uline{39.7} & \uline{44.0} &\uline{37.9}  \\ 
      \textbf{with FLEURS Data}  && & && \\ 
      \rowcolor{Gray}{LLM-SRT-3B} &52&{24.9} &37.3  &40.7 & 34.3 \\
            \rowcolor{Gray}{LLM-SRT-7B} &52&{26.6} &40.1  &41.4 & 36.0 \\
      \rowcolor{Gray}{LLM-SRT-32B} &52&{30.2} &\textbf{43.1}  &43.5 & \textbf{38.9} \\
      \textbf{with CoVoST-2 Data}  && & && \\ 
      \rowcolor{Gray}{Baseline-3B$^\ast$}&430&23.6&34.7&38.6&32.3\\
      \rowcolor{Gray}{Baseline-7B$^\ast$}&430&25.3&36.5&40.2&34.1\\
         \rowcolor{Gray}{LLM-SRT-3B$^\ast$} &430&{26.5} &39.4  &44.0 & 36.6 \\
             \rowcolor{Gray}{LLM-SRT-7B}$^\ast$ &430&{28.7} &\textbf{41.6}  &\textbf{47.1} & \textbf{39.1}\\ \bottomrule

  \end{tabular}

  \caption{\textbf{BLEU Scores on CoVoST-2.} 
* indicates trained on CoVoST-2 dataset.}
  \label{tab:covost2}
\end{table}
\paragraph{Eng$\rightarrow$X S2TT.} As shown in Table \ref{tab:fleurs_15}, the relatively weaker performance of our model in Eng$\rightarrow$X translations is primarily due to the limited amount of English data, stemming from the balanced FLEURS dataset, which contains fewer than 10 hours of data per language. Consequently, as shown in Table \ref{tab:covost2}, our 3B model underperforms compared to SeamlessM4T-V2, but our method is more efficient in data-constrained scenarios.

\paragraph{Scaling Law of S2TT Data.} As shown in Table \ref{tab:covost2}, we compare the performance of Eng$\rightarrow$X translations in both low-resource and high-resource scenarios, with data scales ranging from 52 hours to 430 hours. The results indicate that both our 3B (34.3$\rightarrow$36.6) and 7B (35.0$\rightarrow$39.1) models show consistent performance improvements as the data scale increases, demonstrating that our approach is effective in low-resource settings and scales well with more data.

\paragraph{MLLM Architecture.}  
Our Baseline-7B model follows the Qwen2-Audio setup but keeps the speech encoder frozen, leading to lower performance (34.1 vs. 37.9).

LLM-SRT-7B employs a curriculum learning strategy that incrementally fine-tunes only the adapter layer, allowing the model to adapt more effectively while maintaining efficiency. This approach leads to notable improvements in performance by enabling stable and gradual optimization. Although unfreezing both the speech encoder and the LLM could yield further performance gains, it would also result in significantly higher computational costs.

\paragraph{Multi-task Model LLM-SRT.}  
LLM-SRT is a multi-task model that supports ASR, SMT, and SRT tasks. As shown in Table \ref{tab:mmt}, we evaluated the performance of all tasks.

We found that performing the SRT task did not degrade, but rather slightly improved, ASR performance. Moreover, with the correct transcription, the model achieved a high BLEU score in the SMT task.

\begin{table}[H]
  \centering
    \small
  \begin{tabular}{cccccc} \toprule 
  \textbf{Model}&\textbf{Task} &\textbf{WER$\downarrow$}&Deu&Jpn&Zho \\ \hline
            \multirow{3}{*}{\shortstack{LLM-SRT-7B$^\ast$ \\ Eng$\rightarrow$X}}&ASR& 11.1 &   - &-&- \\
           &SMT& 0 & 32.8&43.6&55.6 \\
           &SRT& 10.9 &28.7 &41.6&  47.1  \\ \bottomrule

  \end{tabular}

\caption{\textbf{Performance of Different Tasks.} 
We evaluated the model on the CoVoST-2 dataset and found that the SMT task, which uses both speech and ground-truth transcription as inputs, achieved a notably high BLEU score.}

  \label{tab:mmt}
\end{table}

\paragraph{Inference Speed.}

As shown in Table \ref{tab:speed}, our method achieves nearly a 3x improvement in inference speed compared to the Qwen2-Audio model, which has a similar parameter size. Even with beam search enabled (5 beams) in our model, the speed remains faster than the greedy search of Qwen2-Audio.

The speed difference is mainly due to our optimized speech adapter design, which compresses the audio features to a fixed size of 80, significantly reducing the token length input to the LLM and accelerating inference.

 \begin{table}[htb]
  \centering
    \small
    \renewcommand{\arraystretch}{0.9}
  \begin{tabular}{cccc} \toprule 
    {\textbf{Model}}&\textbf{Strategy}& \textbf{Batch}&\textbf{Time(s)$\downarrow$}\\ \midrule

 \multirow{3}{*}{Qwen2-Audio}&\multirow{2}{*}{Greedy Search}& 4&59 \\
  & & 8& / \\  \cmidrule(r){2-4}
  &Beam Search 5& 4& / \\ \midrule
   \multirow{8}{*}{LLM-SRT-7B$^\ast$}&\multirow{5}{*}{Greedy Search}& 4&74 \\
    && 8& 39 \\ 
  && 16& 28 \\ 
  && 32& 22 \\ 
    && 64& \textbf{19} \\     \cmidrule(r){2-4}
  &\multirow{3}{*}{Beam Search 5}& 4&93  \\
    && 8& 64 \\ 
    && 12& 56 \\ \bottomrule

  \end{tabular}
\caption{\textbf{Inference Speed Comparison.} We compared the inference time for processing 1,000 speech samples between Qwen2-Audio and LLM-SRT-7B, both with similar parameter sizes, using a 4-card 4090 DDP FP16 inference setup. LLM-SRT-7B demonstrated a 3x speed improvement. / Indicates out-of-memory issues.}

  \label{tab:speed}
\end{table}
\subsection{Ablation Study}
\paragraph{Effect of the ASR, SMT, and SRT Tasks.}
Initially, ASR training was performed to establish a strong baseline~\cite{bansal2018pre, le2023pre}. Removing ASR pre-training resulted in a 2.3-point decrease in BLEU score, emphasizing its importance in S2TT. This section explores the effect of skipping the SMT task and proceeding directly to SRT. As shown in Table~\ref{tab:ablation}, removing SMT and SRT resulted in performance drops of 1.1 and 4.9 points, respectively. While direct SRT maintains an MT-S2TT link with only minor degradation, omitting SRT and relying solely on instruction fine-tuning leads to a substantial performance drop.

\begin{table}[htbp]
  \centering
  \renewcommand{\arraystretch}{1}  
        \setlength{\tabcolsep}{2pt} 

    \small
  \begin{tabular}{lccccccc} \toprule 
  \textbf{Model}&Deu&Jpn&Zho& Avg. \\ \hline

          LLM-SRT-7B$^\ast$  & 28.7 &41.6 &47.1  & 39.1\\
        \hspace{1em}w/o ASR   &26.4{\footnotesize(\textcolor{Red}{-2.3})} &38.6{\footnotesize(\textcolor{Red}{-3.0})}&45.5{\footnotesize(\textcolor{Red}{-1.6})} & 36.8{\footnotesize(\textcolor{Red}{-2.3})} \\
         \hspace{1em}w/o SMT & 27.6{\footnotesize(\textcolor{Red}{-1.1})}&39.7{\footnotesize(\textcolor{Red}{-1.9})}&46.5{\footnotesize(\textcolor{Red}{-0.6})} &38.0{\footnotesize(\textcolor{Red}{-1.1})}  \\
          \hspace{1em}w/o SRT &25.6{\footnotesize(\textcolor{Red}{-3.1})}&36.7{\footnotesize(\textcolor{Red}{-4.9})}  & 40.4{\footnotesize(\textcolor{Red}{-6.7})}  &34.2{\footnotesize(\textcolor{Red}{-4.9})}   \\
\bottomrule
  \end{tabular}

  \caption{\textbf{Ablation Study.} We evaluated the model on the CoVoST-2 Eng$\rightarrow$X dataset. Removing the SRT task leads to a substantial reduction in BLEU score, underscoring its critical role in overall performance.
}
  \label{tab:ablation}
\end{table}

\begingroup
\renewcommand{\arraystretch}{1.15} 
\begin{CJK*}{UTF8}{gbsn}
\begin{table*}[t]
    \small 
    \centering
      \setlength{\tabcolsep}{0pt} 
    \setlength{\extrarowheight}{1pt}
    \resizebox{1\textwidth}{!}{\begin{tabular}{lll}
    \toprule

    \textbf{Model}& &\textbf{BLEU}$\uparrow$ \\ \midrule

    \tiny \textbf{Audio}     &\hspace{0.2em}\tiny三国是中国古代历史上最血腥的时代之一。成千上万的人为了争夺西安豪华宫殿最高的权力而死去。 &\\
    \tiny \textbf{Ground-truth}& \textcolor{blue}{\begin{CJK*}{UTF8}{min} \tiny 三国志は、古代中国の歴史の中で最も血なまぐさい時代の1つでした。西安の大宮殿の最高位を狙う争いの中で何千人もが戦士しました。 \end{CJK*}}&\\
   \cmidrule(r){1-2}
    
    \tiny \textbf{SeamlessM4T-V2}  &\textcolor{purple}{\begin{CJK*}{UTF8}{min} \tiny三国は中国古代歴史で最も血腥な時代の一つ 千上万の人が シアン豪華宮殿の最高権力を争うために死ぬ\end{CJK*}} huan& \multicolumn{1}{l}{14.7~\textbar~\textit{5.6}}\\

      \tiny \textbf{Qwen2-Audio}  &\textcolor{purple}{\begin{CJK*}{UTF8}{min} \tiny三国は中国の歴史で最も血なましい時代の一つです。何千人もの人が、西安の豪華宮殿の権力を得るために死んでいます。   \end{CJK*}} &27.9~\textbar~\textit{12.5}  \\ \cmidrule(r){2-2}

   \tiny \multirow{2}{*}{\textbf{LLM-SRT-3B}}  &{\tiny\hspace{0.2em}三国是中国古代历史上最血腥的时期之一，成千上万人为了争夺西安豪华宫殿的最高权力而死去。\texttt{<|zho|><|jpn|>}}         \\
   &\textcolor{purple}{\begin{CJK*}{UTF8}{min} \tiny三国時代は中国の歴史上で最も血なまぐさい時代の一つで、西安の豪華な宮殿で最高の権力を争うために、数万人が死にました。 \end{CJK*}}&34.9~\textbar~\textit{16.5}\\

   \tiny \multirow{2}{*}{\textbf{LLM-SRT-32B}}  &{\tiny\hspace{0.2em}三国是中国古代历史上最血腥的时代之一。成千上万的人为了争夺西安豪华宫殿的最高权力而死去。\texttt{<|zho|><|jpn|>}}        \\
   &\textcolor{purple}{\begin{CJK*}{UTF8}{min} \tiny三国時代は、中国の歴史の中で最も血なまぐさい時代の1つで、西安の豪華な宮殿の支配権を巡って何千人もの人々が死にました。 \end{CJK*}}&49.3~\textbar~\textit{40.4}\\

          \bottomrule

\end{tabular}}
  \caption{\textbf{Case Study.} We compare the BLEU scores of our method with those of other approaches, using the 'char' tokenizer (denoted in regular font) and the 'ja-mecab' tokenizer (presented in \textit{italics}). With a transcription Character Error Rate (CER) score of 11.4 for the 3B model and 4.6 for the 32B model.  }
    \label{tab:case_study}
\end{table*}
\end{CJK*}
\endgroup

\paragraph{Case Study.}
As shown in Tables \ref{tab:case_study} and \ref{tab:case_study2}, SeamlessM4T-V2 demonstrates poor performance in Japanese, achieving the lowest BLEU score. Qwen-Audio outperforms it, while our method outperforms significantly. Our approach follows a two-step process: first, it generates a transcription of the input speech, which is then used to produce the translation. This strategy ensures that the translation benefits from the transcribed text, leveraging its structure and context to improve accuracy. By incorporating transcription into the translation process, our method minimizes ambiguities and enhances translation quality, especially in complex or context-dependent scenarios.

\section{Related Work}

\paragraph{Cascaded S2TT.} This method follows a two-step process: first, ASR transcribes the spoken language into text, and then MT translates the transcribed text into the target language. This approach leverages the strengths of specialized ASR~\cite{radford2023robust,baevski2020wav2vec} and MT~\cite{fan2021beyond} models, utilizing extensive training data and advanced techniques. ASR models accurately convert speech to text, while sophisticated MT models, benefiting from large multilingual datasets, translate with high accuracy and fluency. However, the cascaded approach is prone to error propagation.

\paragraph{End-to-End S2TT.} In this paradigm, a single model is trained to directly map speech from the source language to text in the target language, skipping the intermediate transcription step~\cite{wang2020improving,wang2020fairseq,inaguma2020espnet}. Early work on joint speech recognition and translation primarily used streaming models, aiming to provide real-time multilingual synchronization~\cite{sperber2020consistent,dong2021consecutive,papi2024leveraging}. These pioneering efforts focused more on reducing latency and enhancing efficiency than offline speech translation systems. End-to-end ST offers several advantages, including reduced latency, simplified system architecture, and the elimination of error propagation between the ASR and MT stages. Despite these benefits, end-to-end ST models face challenges, such as the need for extensive parallel speech-to-text data, which is resource-intensive and difficult to obtain.

\paragraph{Audio MLLMs.} Recent advancements in audio MLLMs~\cite{li2025perception} have significantly improved speech recognition and translation. SpeechGPT~\cite{zhang2023speechgpt} uses prompting to enhance speech recognition in large language models. BLSP-KD~\cite{wang2024blsp} refines speech-text alignment through knowledge distillation. SALMONN~\cite{tang2023salmonn} aims to improve auditory comprehension of language and music in models. Qwen-Audio~\cite{Qwen-Audio} advances audio recognition and translation by retraining speech encoders within a multi-task framework. Qwen2.5-Omni~\cite{Qwen2.5-Omni}, the end-to-end multimodal model, is designed for comprehensive multimodal perception, seamlessly  processing heterogeneous input modalities.  

\section{Conclusion}

In this paper, we propose a novel strategy that reformulates the speech-to-text translation task as a combination of speech recognition and translation tasks, leveraging the machine translation capabilities of LLMs to enhance the performance of MLLMs in S2TT. To validate our approach, we train three MLLMs with sizes 3B, 7B, and 32B, and implement a three-stage curriculum learning strategy, which proves highly effective in low-resource scenarios while further improving performance when sufficient training data is available. Our model achieves state-of-the-art results across 15×14 translation directions, excelling in low-resource learning on the FLEURS dataset and supervised training on the CoVoST-2 dataset. These results highlight the robustness and effectiveness of our approach across diverse linguistic and data availability settings.

For future work, we aim to further optimize the LLM-SRT model to push its performance boundaries and extend its application to a broader range of languages.

\section*{Limitations}
This paper presents a method for training an MLLM for languages with less than 10 hours of speech translation data.

However, the performance of S2TT and the range of supported languages are constrained by the capabilities of the LLM. MLLMs trained using this method may not perform well on languages that are not supported by the LLM or on those with poor machine translation performance.

\section*{Acknowledgement}

The research in this article is supported by the National Natural Science Foundation of China (Grants No. U22B2059 and 62276083) and the National Science and Technology Innovation 2030 Major Program (Grant No. 2024ZD01NL00101). We also appreciate the support from China Mobile Group Heilongjiang Co., Ltd. @ on our research, the research is a joint collaboration between the involved parties.

\normalem
\bibliography{anthology,custom}

\appendix
\section{Appendix}
In this chapter, we report the language support in Table \ref{tab:all_languages}, the case study in Table \ref{tab:case_study2}, and the BLEU scores for the 15x14 directions of the FLEURS dataset in Table \ref{fleurs_15x14}.

\begin{table*}[h]  
\centering
\begin{tabular}{lllll}
\toprule
\textbf{Code} & \textbf{Language} & \textbf{Family} & \textbf{Subgrouping} & \textbf{Script}  \\
\midrule
deu & German & Indo-European & Germanic & Latn \\
eng & English & Indo-European & Germanic & Latn  \\
fra & French & Indo-European & Italic & Latn \\
ind & Indonesian & Austronesian & Malayo-Polynesian & Latn \\

ita & Italian & Indo-European & Italic & Latn  \\
jpn & Japanese & Japonic & Japanesic & Jpan  \\
kor & Korean & Koreanic & Korean & Kore   \\
nld & Dutch & Indo-European & Germanic & Latn   \\
por & Portuguese & Indo-European & Italic & Latn   \\
rus & Russian & Indo-European & Balto-Slavic & Cyrl   \\
spa & Spanish & Indo-European & Italic & Latn  \\
tha & Thai & Tai-Kadai & Kam-Tai & Thai  \\
vie & Vietnamese & Austroasiatic& Vietic & Latn  \\
yue & Cantonese & Sino-Tibetan & Sinitic & Hant  \\
zho & Chinese & Sino-Tibetan & Sinitic & Hans  \\
\bottomrule
\end{tabular}
\caption{
\textbf{Language Support.} The table lists language codes, names, families, subgroups, and scripts. Language support is extensible based on ISO 639-3. 6$\times$12 translation directions cover source languages (eng, deu, fra, zho, rus, jpn) and target languages (eng, deu, fra, spa, por, ita, nld, rus, jpn, kor, vie, ind, zho).
}
\label{tab:all_languages}

\end{table*}

\begin{table*}[h]
  \centering
    \small
  \begin{tabular}{ccccccccc} \toprule 
\textbf{Stage} &\textbf{Dataset}&\textbf{Hour}&\textbf{Batch}&\textbf{Step}&\textbf{Learning Rate}&\textbf{Warmup Step}&\textbf{Optimizer} \\ \hline
            \multirow{2}{*}{ASR}& Common Voice 19  & 4498 & \multirow{4}{*}{16} &\multirow{2}{*}{472000}&{1e-4}&\multirow{4}{*}{1000}&\multirow{4}{*}{AdamW}\\
            & FLEURS&  {129}&&&{1e-4}\\
           SMT&  FLEURS&129 & &44000&{1e-4}\\
           SRT&  FLEURS&129 & &83000&1e-5\\ \bottomrule

  \end{tabular}

\caption{\textbf{Training Details for LLM-SRT-3B-V2.} The step count refers to the number of steps on a single GPU.}

  \label{tab:details}
\end{table*}

\begin{table*}[h]
  \centering
    \small
  \begin{tabular}{ccccccccc} \toprule 
\textbf{Model} &\textbf{Encoder}&\textbf{Adapter}&\textbf{LLM}&\textbf{Language Support}\\ \hline
            LLM-SRT-3B& & \multirow{4}{*}{Q-Former + MLP} &Qwen2.5-3B&{$6\times12$} \\
           LLM-SRT-7B& Whisper large-v3’s & &Qwen2.5-7B &{$6\times12$}\\
           LLM-SRT-32B& encoder & & Qwen2.5-32B&{$6\times12$}\\ 
                       LLM-SRT-3B-V2&  & &Qwen2.5-3B &$15\times14$ \\

           \bottomrule

  \end{tabular}

\caption{\textbf{Model Settings.} The V2 model employs the same architecture but supports more languages.}

  \label{tab:models}
\end{table*}

\begin{table*}[h]
    \centering
    \footnotesize
    \label{tab:lora_settings}
    \renewcommand{\arraystretch}{1.2}
    \setlength{\tabcolsep}{3mm}
    \begin{tabular}{lccccc}
        \toprule
        \textbf{Model} & \textbf{Rank (r)} & \textbf{Alpha} & \textbf{Dropout} & \textbf{Target Keys} & \textbf{Bias} \\
        \midrule 
        SeamlessM4T-V2 & 32 & 64 & 0.2 & \texttt{.*\_proj} & \texttt{none} \\
        LLM-SRT-3B-V2 & 8 & 32 & 0.05 & \texttt{q\_proj, v\_proj} & \texttt{none} \\
        \bottomrule
    \end{tabular}
        \caption{\textbf{LoRA Configuration Settings.} We only fine-tuned the LLM with LoRA for LLM-SRT-3B-V2.}
        \label{tab:lora}
\end{table*}

\begin{table*}[htbp]
    \centering
    \footnotesize
    \label{tab:parameters}
    \renewcommand{\arraystretch}{1.2}
    \setlength{\tabcolsep}{2mm}
    \begin{tabular}{lrcl}
        \toprule
        \multirow{2}{*}{\textbf{Modules}} & \multirow{2}{*}{\textbf{Param}} & \textbf{Training} & \multirow{2}{*}{\textbf{Details}}  \\
        & & \textbf{stage} & \\
        \midrule 
        Speech Encoder & $\sim$635M & -- & Whisper's encoder  \\
        \midrule
        Speech Adapter & {\color{red}$\sim$73.7M} & I\&II\&III & Q-Former and MLP \\
        \midrule
        LLM & $\sim$3.1B & -- & Qwen2.5-3B \\
        \midrule
        LLM adapter & {\color{red}$\sim$1.8M} & III & LoRA \\
        \midrule
        Total & $\sim$3.8B & & \\
        \bottomrule
    \end{tabular}
    \caption{\textbf{LLM-SRT-3B-V2 Training Parameters.} The {\color{red}red} indicates trainable parameters.}
    \label{tab:train}
\end{table*}

\begin{table*}[htbp]
    \centering
    \footnotesize
    \renewcommand{\arraystretch}{1.2}
    \setlength{\tabcolsep}{2mm}
    \begin{tabular}{ccccccccc}
        \toprule
\textbf{X$\rightarrow$14}	&	\textbf{SeamlessM4T-V2}	&	\textbf{SeamlessM4T-V2+Lora}	&	\textbf{LLM-SRT-3B}	&	\textbf{LLM-SRT-3B+Lora}	\\ \midrule
deu	&	21.6 / 76.2	&	22.0 / 76.8	&	23.4 / 83.4	&	24.5 / 83.5	\\
eng	&	33.3 / 84.8	&	32.8 / 84.3	&	27.8 / 85.5	&	29.1 / 85.4	\\
fra	&	21.1 / 77.6	&	21.3 / 78.0	&	24.0 / 83.5	&	24.3 / 83.6	\\
ind	&	18.4 / 73.8	&	19.1 / 75.0	&	22.4 / 82.3	&	22.9 / 82.2	\\
ita	&	19.1 / 76.9	&	19.4 / 77.4	&	22.0 / 83.5	&	22.8 / 83.5	\\
jpn	&	14.5 / 72.4	&	14.4 / 72.5	&	15.3 / 80.1	&	16.5 / 80.3	\\
kor	&	17.2 / 76.6	&	17.0 / 76.5	&	17.9 / 81.8	&	18.0 / 81.9	\\
nld	&	18.4 / 74.8	&	18.6 / 75.3	&	20.4 / 81.3	&	20.9 / 81.4	\\
por	&	18.8 / 73.3	&	18.6 / 73.5	&	25.7 / 84.1	&	26.2 / 84.2	\\
rus	&	20.6 / 75.9	&	21.3 / 76.7	&	22.8 / 82.5	&	23.3 / 82.5	\\
spa	&	17.6 / 76.1	&	18.1 / 76.9	&	21.9 / 83.9	&	22.6 / 83.9	\\
tha	&	12.8 / 71.1	&	13.2 / 71.7	&	12.5 / 78.6	&	14.7 / 78.8	\\
vie	&	17.5 / 74.1	&	17.5 / 74.1	&	18.2 / 79.7	&	19.0 / 79.8	\\
yue	&	15.2 / 70.9	&	14.9 / 70.5	&	15.5 / 78.9	&	16.3 / 79.0	\\
zho	&	16.8 / 77.5	&	17.0 / 77.7	&	19.2 / 82.5	&	19.6 / 82.6	\\ \midrule
Avg.	&	18.8 / 75.5	&	19.0 / 75.8	&	20.6 / 82.1	&	21.4 / 82.2	\\
        \bottomrule
    \end{tabular}
    \caption{\textbf{BLEU/COMET Scores for 15 Languages Across 15×14 Translation Directions on FLEUR.} Detailed results are summarized in Tables~\ref{fleurs_15x14}.}
    \label{tab:comet}
\end{table*}

\begin{CJK*}{UTF8}{gbsn}
\begin{table*}[h]  
    \renewcommand{\arraystretch}{1.2} 
    \centering
    \setlength{\tabcolsep}{4pt} 
    \setlength{\extrarowheight}{1pt}
    \resizebox{1\textwidth}{!}{
    \begin{tabular}{llc}
    \toprule
    \textbf{Case} &  & \textbf{BLEU}$\uparrow$ \\ \midrule

    \multirow{2}{*}{\textbf{Audio}} & \hspace{0.2em}There may be more maria on the near side because the crust is thinner. It was easier for lava to rise up to the surface. &  \\
     & \textcolor{blue}{ 地殻が薄いため、近い方には海が多くなることがあります。溶岩が浮上しやすくなっていました。
} &  \\ \cmidrule(r){1-2}

    \textbf{Qwen2.5-3B}~(MT) & \textcolor{purple}{ 	近側には可能にマリアがあるかもしれません because the crustははがより薄いからマagmaが表面に上りやすかった lavaです。} & 2.9 \\

    \multirow{1}{*}{\textbf{LLM-SRT-3B}~(S2TT)}& \textcolor{purple}{近側にはもっとマリアがある可能性があります。地殻が薄いためです。溶岩が表面に上りやすかったからです。} & 30.6 \\

    \bottomrule

    \end{tabular}}
     \caption{\textbf{Error Analysis.} We observe that the Qwen2.5-3B model exhibits language mixing (e.g., Japanese and Cantonese) in certain translation directions. Consequently, S2TT outperforms MT in these scenarios.}  
    \label{tab:error}
\end{table*}
\end{CJK*}

\begin{CJK*}{UTF8}{gbsn}
\begin{table*}[h]  
    \renewcommand{\arraystretch}{1.2} 
    \centering
    \setlength{\tabcolsep}{4pt} 
    \setlength{\extrarowheight}{1pt}
    \resizebox{1\textwidth}{!}{
    \begin{tabular}{llc}
    \toprule
    \textbf{Case} &  & \textbf{BLEU}$\uparrow$ \\ \midrule

    \multirow{2}{*}{\textbf{Audio \# 1}} & \hspace{0.2em}它让玩家可以通过在空中移动设备来控制电子游戏中的运动和操作。 &  \\
     & \textcolor{blue}{ This will allow players to control actions and movements in video games by moving the device through the air.} &  \\ \cmidrule(r){1-2}

    \textbf{SeamlessM4T-V2} & \textcolor{purple}{ It allows players to control the movement and operation of electronic games through mobile devices in the air.} & 14.0 \\

    \multirow{2}{*}{\textbf{LLM-SRT-3B}} & \hspace{0.2em}它让玩家可以通过在空中移动设备来控制电子游戏中的动作和操作。\texttt{<|zho|><|eng|>} &  \\
    & \textcolor{purple}{ It allows players to control the actions and operations of an electronic game by moving the device in the air.} & 23.2 \\

    \multirow{2}{*}{\textbf{LLM-SRT-32B}} & \hspace{0.2em}它让玩家可以通过在空中移动设备来控制电子游戏中的动作和操作。 \texttt{<|zho|><|eng|>}\\
    & \textcolor{purple}{It allows players to control actions and operations in electronic games by moving the device in the air.} & 43.0 \\
\bottomrule

    \end{tabular}}
     \caption{\textbf{Case Study.} We compare the BLEU scores of our models with SeamlessM4T-V2.}
    \label{tab:case_study2}
\end{table*}
\end{CJK*}

\clearpage  
\begin{table}[t]
\tiny
\vspace*{-30pt} 
\centering
\renewcommand{\arraystretch}{0.8}
\setlength{\tabcolsep}{5pt} 
\begin{tabular}{lllllllll}
\toprule
\multirow{2}{*}{src\_tgt} & \multicolumn{4}{c}{\textbf{SeamlessM4T-V2}}                              & \multicolumn{4}{c}{\textbf{LLM-SRT-3B-V2}}                               \\
                          & BLEU & \multicolumn{1}{l}{Avg.} & +Lora & Avg.                                       & BLEU                 & Avg.                                       & +Lora    & \multicolumn{1}{l}{Avg.}     \\ \midrule
deu\_eng                  & 37.1 & \multirow{14}{*}{21.6}   & 36.9 & {\multirow{14}{*}{22.0}} & 32.3                 & {\multirow{14}{*}{23.4}} & 34.3    & \multirow{14}{*}{24.5}       \\
deu\_fra                  & 19.3 &                          & 21.2 & {}                       & 23.2                 & {}                       & 25.0    &                              \\
deu\_ind                  & 19.1 &                          & 19.4 & {}                       & 17.2                 & {}                       & 18.8    &                              \\
deu\_ita                  & 12.5 &                          & 13.5 & {}                       & 13.8                 & {}                       & 14.8    &                              \\
deu\_jpn                  & 26.9 &                          & 25.8 & {}                       & 30.2                 & {}                       & 31.9    &                              \\
deu\_kor                  & 29.1 &                          & 29.0 & {}                       & 23.1                 & {}                       & 24.6    &                              \\
\bottomrule
\end{tabular}
\end{table}

\begin{table}[!p]
\tiny
\centering
\vspace*{-71pt} 
\renewcommand{\arraystretch}{0.8}
\setlength{\tabcolsep}{5pt} 
\begin{tabular}{lcccccccclllllllllll}
\toprule
\multirow{2}{*}{src\_tgt} & \multicolumn{4}{c}{\textbf{SeamlessM4T-V2}}                              & \multicolumn{4}{c}{\textbf{LLM-SRT-3B-V2}}                               \\
                          & BLEU & \multicolumn{1}{l}{Avg.} & +Lora & Avg.                                       & BLEU                 & Avg.                                       & +Lora    & \multicolumn{1}{l}{Avg.}     \\ \midrule
por\_deu & 6.8  & \multirow{14}{*}{18.8} & 7.5  & \multicolumn{1}{c}{\multirow{14}{*}{18.6}} & 18.2                 & \multicolumn{1}{c}{\multirow{14}{*}{25.7}} & 18.5 & \multirow{14}{*}{26.2}  \\
por\_eng & 38.4 &                        & 37.7 & \multicolumn{1}{c}{}                       & 38.3                 & \multicolumn{1}{c}{}                       & 40.8 &                         \\
por\_fra & 16.3 &                        & 16.7 & \multicolumn{1}{c}{}                       & 28.8                 & \multicolumn{1}{c}{}                       & 29.6 &                         \\
por\_ind & 13.9 &                        & 15.2 & \multicolumn{1}{c}{}                       & 20.5                 & \multicolumn{1}{c}{}                       & 21.6 &                         \\
por\_ita & 10.3 &                        & 10.4 & \multicolumn{1}{c}{}                       & 17.4                 & \multicolumn{1}{c}{}                       & 17.5 &                         \\
por\_jpn & 24.5 &                        & 23.4 & \multicolumn{1}{c}{}                       & 31.8                 & \multicolumn{1}{c}{}                       & 32.8 &                         \\
\bottomrule
\end{tabular}
\caption{\textbf{BLEU Scores for 15 Languages Across 15×14 Translation Directions on FLEURS.}}

\label{fleurs_15x14}
\end{table}

\end{document}